\documentclass[conference]{IEEEtran}
\IEEEoverridecommandlockouts
\usepackage{cite}
\usepackage{amsmath,amssymb,amsfonts}
\usepackage{algorithm}
\usepackage{algpseudocode} 
\usepackage{graphicx}
\usepackage{array}
\usepackage{textcomp}
\usepackage{booktabs}
\usepackage{multirow}
\usepackage{caption}
\usepackage{subcaption}
\usepackage{makecell}
\usepackage{tikz}
\usepackage{comment}
\usepackage{url}
\usetikzlibrary{positioning,arrows.meta}
\def\BibTeX{{\rm B\kern-.05em{\sc i\kern-.025em b}\kern-.08em
    T\kern-.1667em\lower.7ex\hbox{E}\kern-.125emX}}

\usepackage[dvipsnames]{xcolor}
\definecolor{linkblue}{HTML}{3B65AD}
\definecolor{citationblue}{HTML}{3B65AD}
\definecolor{urlblue}{HTML}{2F5597}
\usepackage{hyperref}
\hypersetup{
    colorlinks=true,
    linkcolor=black,
    urlcolor=urlblue,
    citecolor=linkblue,
    pdftitle={A Digital Twin Framework for Traffic-Aware UAV Pavement Monitoring in Open-Traffic Conditions},
    pdfauthor={Yamil Uchani, Grace Luna, Edwin Salcedo, Mauricio Figueroa},
    pdfsubject={Traffic-aware UAV pavement monitoring using a digital twin framework},
    pdfkeywords={Pavement Monitoring, Digital Twin, Traffic-Aware UAV Inspection, Recovery Strategies},
    pdfdisplaydoctitle=true,
    pdfpagemode=UseNone,
    pdfstartview={FitH},
    bookmarksopen=false
}

\definecolor{orcidlogocol}{HTML}{A6CE39}
\DeclareRobustCommand{\orcidicon}{%
\raisebox{-0.2ex}{%
\begin{tikzpicture}[scale=0.13]
\fill[orcidlogocol] (0,0) circle (1);
\node[white, font=\bfseries\tiny] at (0,0) {iD};
\end{tikzpicture}}}
\newcommand{\orcidlinked}[1]{\href{https://orcid.org/#1}{\orcidicon}}

\begin{document}

\title{A Digital Twin Framework for Traffic-Aware UAV Pavement Monitoring in Open-Traffic Conditions}

\author{
\IEEEauthorblockN{
Yamil Uchani\textsuperscript{1}\,\orcidlinked{0009-0001-0624-7341},
Grace Luna\textsuperscript{1,\textdaggerdbl}\,\orcidlinked{0009-0003-5095-8866},
Edwin Salcedo\textsuperscript{2,\textdaggerdbl,*}\,\orcidlinked{0000-0001-8970-8838},
Mauricio Figueroa\textsuperscript{1}\,\orcidlinked{0009-0008-9402-5706}
}
\IEEEauthorblockA{
\textsuperscript{1}\textit{Department of Mechatronics Engineering, Universidad Cat\'olica Boliviana ``San Pablo''}, Bolivia\\
\textsuperscript{2}\textit{School of Electronic Engineering and Computer Science, Queen Mary University of London}, United Kingdom\\
\textsuperscript{\textdaggerdbl}These authors contributed equally to this work.\\
\textsuperscript{*}Corresponding author: e.r.salcedoaliaga@qmul.ac.uk
}
}

\maketitle

\begin{abstract}
UAV-based pavement inspection can reduce the cost and risk of road-surface monitoring, but real-world deployment remains difficult when traffic, pedestrians, and temporary occlusions affect defect visibility. This paper presents a Unity-based digital twin framework for traffic-aware UAV pavement monitoring in open-traffic conditions. The proposed environment integrates procedurally generated road defects, dynamic traffic agents, autonomous UAV navigation, and a multitask YOLOv8n perception module for detecting road defects, pedestrians, and vehicles while classifying road-defect subtypes. After synthetic-domain fine-tuning, the perception model achieved 0.959 mAP@0.5 and 0.940 macro F1-score on a held-out synthetic test set generated from the simulator. The digital twin was then used to evaluate hover-and-recheck, micro-repositioning, and skip-and-revisit recovery strategies across different traffic densities and flight altitudes. Results show that flight altitude strongly affects inspection coverage, while recovery strategies introduce different trade-offs between coverage, mission duration, energy consumption, and revisit behaviour. These findings demonstrate that digital twins can support the development and evaluation of traffic-aware UAV inspection strategies before real-world deployment. The full implementation and trained models are available at \url{https://github.com/EdwinTSalcedo/RDMO-DigitalTwin}.
\end{abstract}

\begin{IEEEkeywords}
Pavement Monitoring, Digital Twin, Traffic-Aware UAV Inspection, Recovery Strategies
\end{IEEEkeywords}

\section{Introduction}
\label{sec:introduction}

Road infrastructure is vital to mobility, economic development, and public safety, yet poor road conditions remain a major cause of accidents, vehicle damage, and costly maintenance worldwide. Road-surface defects such as potholes, cracks, and asphalt delamination can accelerate vehicle wear, degrade ride quality, and create hazardous driving conditions under adverse weather, low-visibility, or high-traffic conditions. For this reason, timely and reliable inspection is essential for identifying road defects and supporting maintenance decisions. Current road inspection approaches rely on manual visual assessment or computer-vision-based inspection pipelines \cite{dhiman2019,salcedo2022}. More recently, the advent of unmanned aerial vehicles (UAVs) and edge computing has enabled greater automation in road inspection tasks \cite{silva2020,zhao2023,lopezgonzalez2026}.

However, despite these advances, existing UAV-based inspection methods remain subject to operational limitations, human error, and safety risks, especially when monitored areas are large or experience high traffic volumes. In such dynamic environments, inspections may require temporary traffic restrictions to ensure road-surface visibility, causing disruption and economic losses. Moreover, road-inspection computer-vision pipelines validated under controlled conditions may perform less reliably when exposed to moving road users and partially occluded defects. Simulation and digital-twin platforms can help reduce this gap by enabling safer, repeatable, and traffic-aware evaluation of UAV inspection strategies before real-world deployment.

Although prior work has explored maintenance-oriented digital twins \cite{sierra2022,talaghat2024,consilvio2023,topu2025} and UAV inspection-route planning \cite{zhao2023,zhong2023}, existing approaches rarely provide a unified traffic-aware testbed that integrates realistic road-surface defect generation, dynamic vehicle and pedestrian flows, UAV inspection behaviour, and recovery strategies for temporarily occluded road regions. This limits the development and validation of UAV-based inspection systems under realistic operating conditions. To address this gap, we present a Unity-based digital-twin framework for UAV-driven road-damage inspection. The framework combines procedural road-defect generation with simulated vehicle and pedestrian traffic, enabling the evaluation of inspection strategies when road regions are temporarily obstructed. Rather than assuming that recovery actions always improve inspection performance, the framework enables evaluation of recovery strategies and reveals their trade-offs in terms of coverage, mission duration, energy consumption, and revisit behaviour. Experiments demonstrate its potential for assessing traffic-aware UAV inspection before deployment.

\begin{figure*}[htpb]
\centering
\begin{minipage}{\textwidth}
\centering

\begin{subfigure}[b]{0.48\linewidth}
\centering
\includegraphics[width=\textwidth]{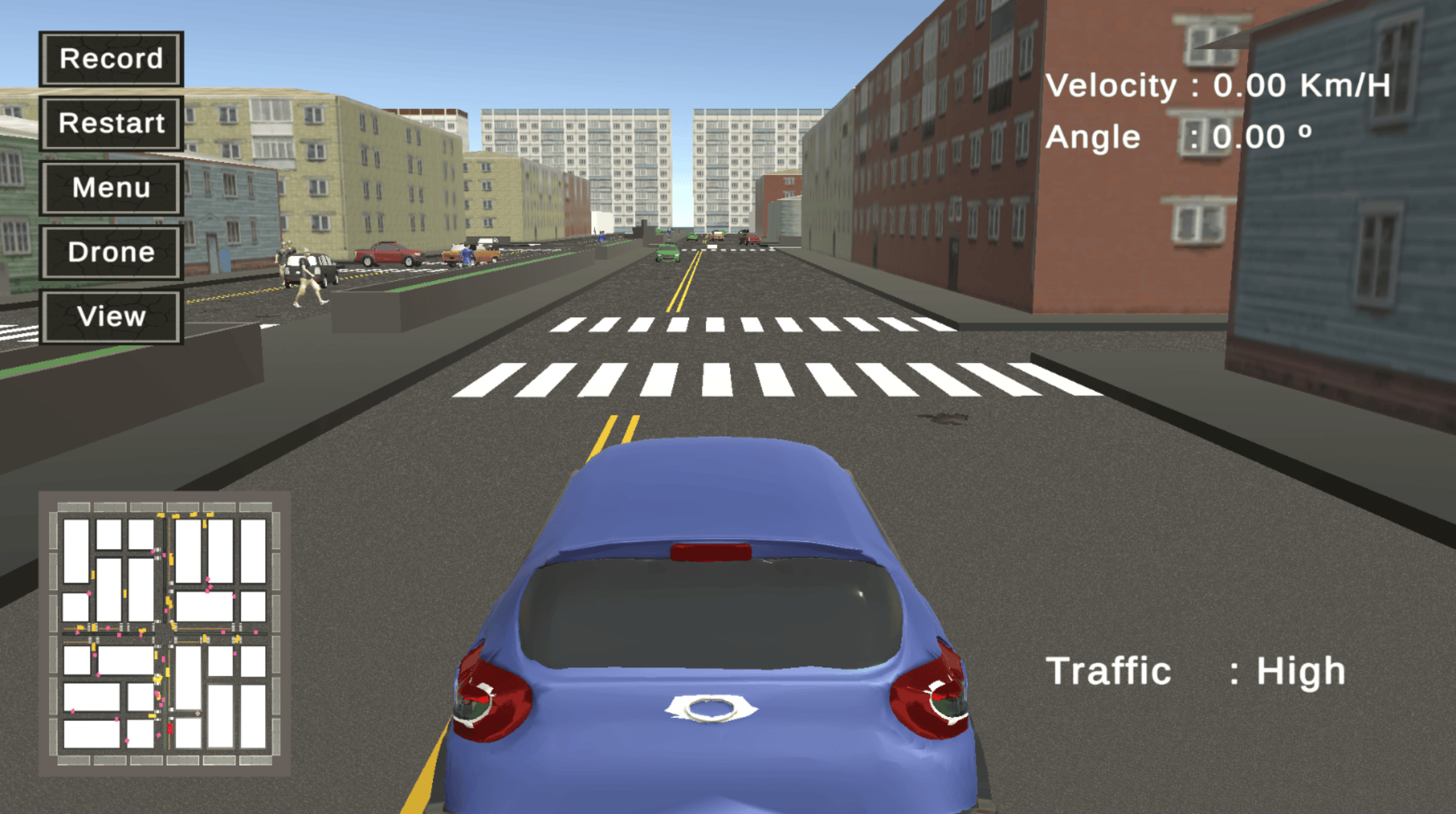}
\caption{}
\label{fig:sim-groundvehicle}
\end{subfigure}
\hfill
\begin{subfigure}[b]{0.48\linewidth}
\centering
\includegraphics[width=\textwidth]{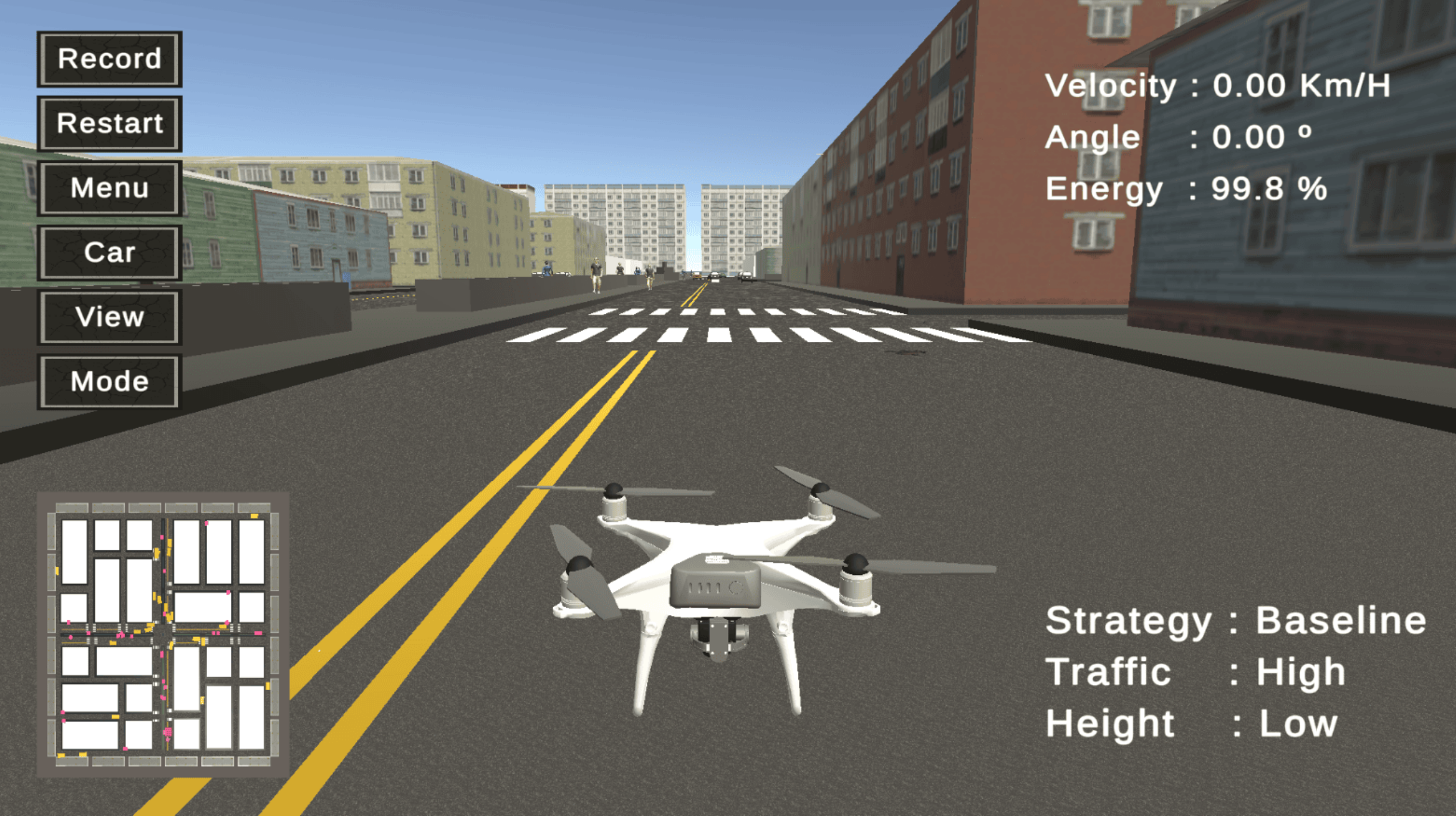}
\caption{}
\label{fig:sim-uav}
\end{subfigure}

\vspace{0.5em}

\begin{subfigure}[b]{0.48\linewidth}
\centering
\includegraphics[width=\textwidth]{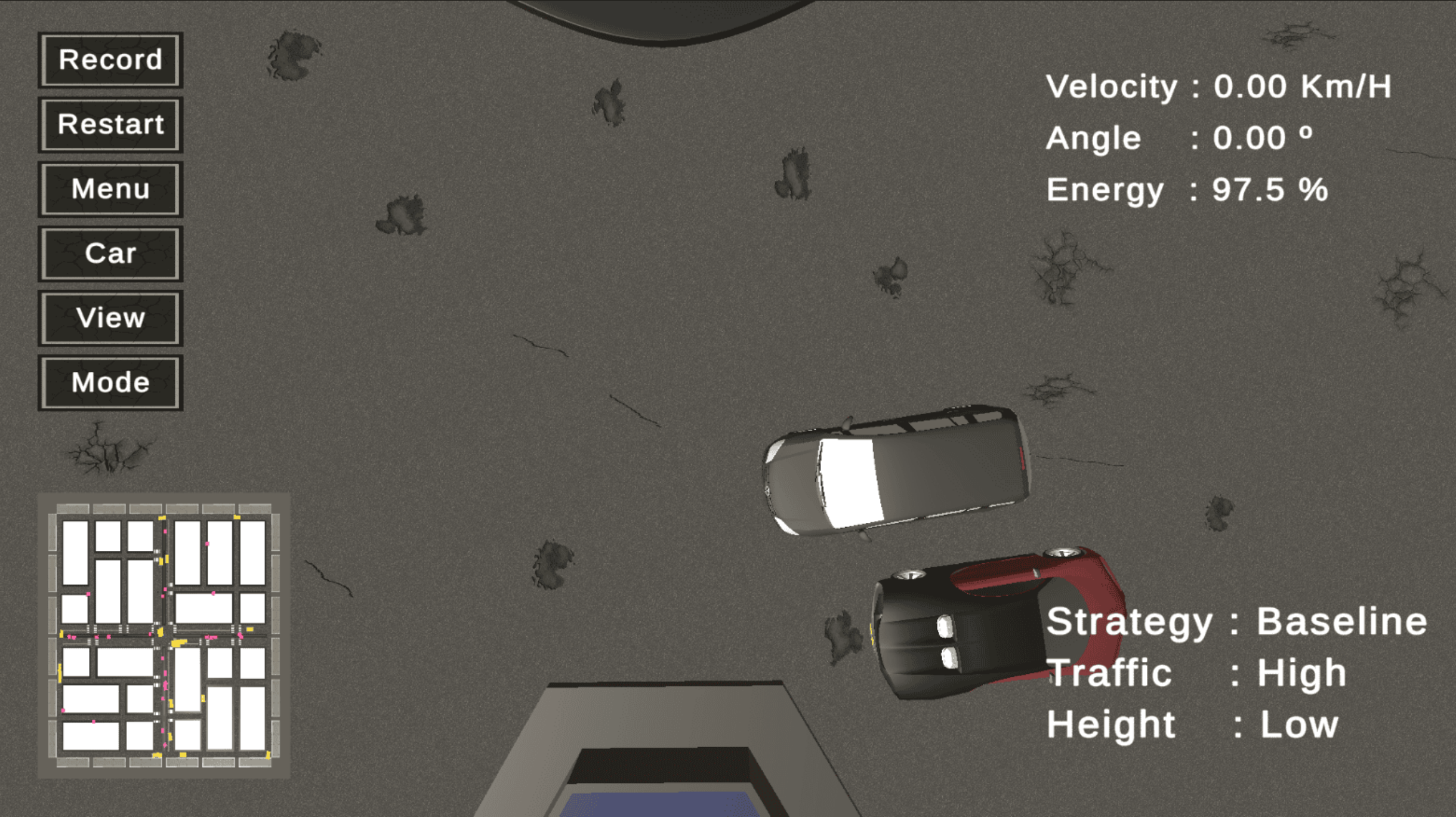}
\caption{}
\label{fig:top-view-uav}
\end{subfigure}
\hfill
\begin{subfigure}[b]{0.48\linewidth}
\centering
\includegraphics[width=\textwidth]{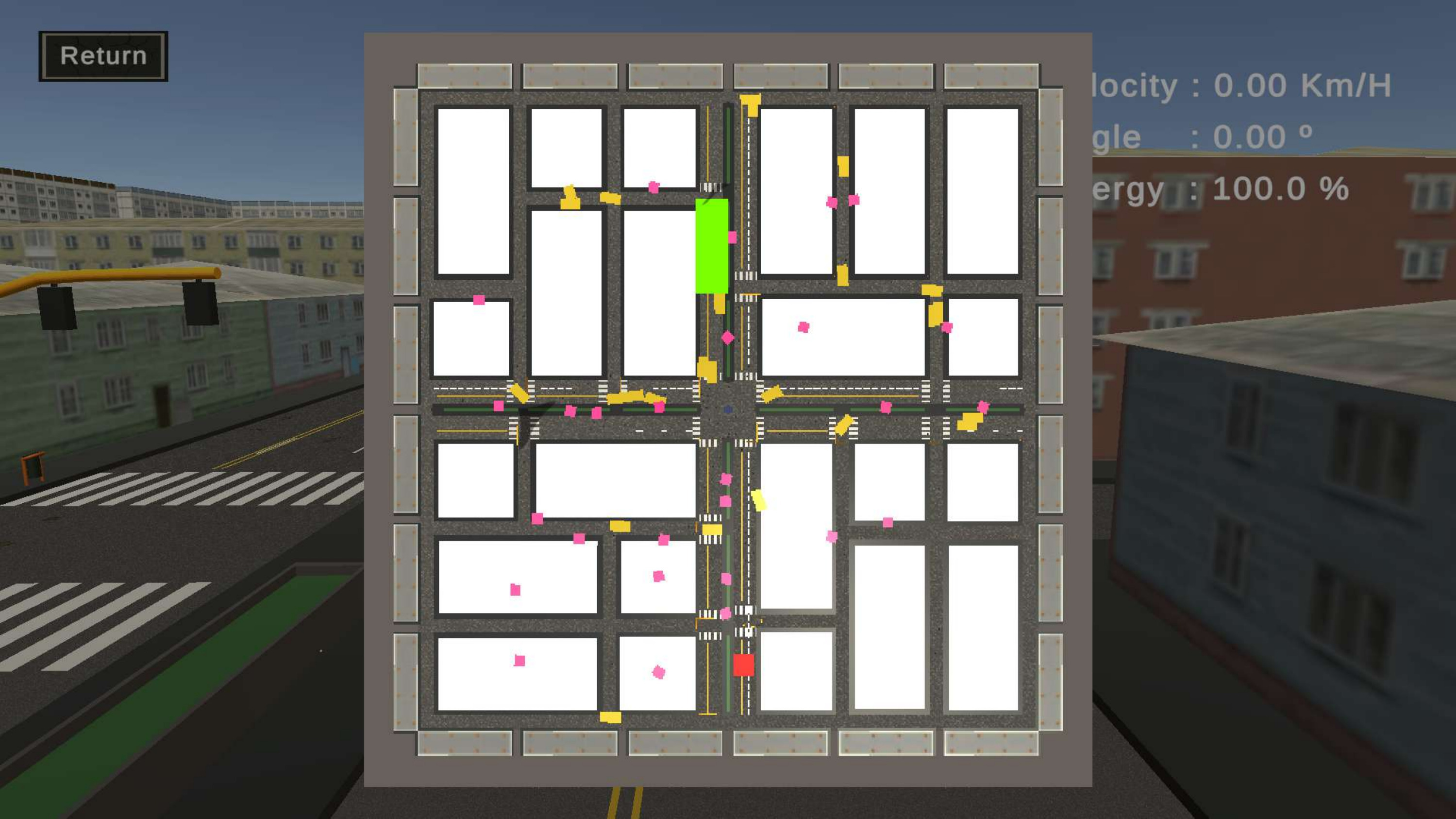}
\caption{}
\label{fig:simulator-map}
\end{subfigure}

\end{minipage}
\caption{Main components of the digital twin simulation framework: (a) ground vehicle, (b) unmanned aerial vehicle (UAV), (c) inspection perspective from the UAV, and (d) map view. The green region denotes a road segment selected for inspection. The red dot represents the UAV, while the yellow and pink dots represent cars and pedestrians, respectively.}
\label{fig:simulator-components}
\end{figure*}

\section{Related Work}
\label{sec:related-work}

UAV-based inspection has become a practical approach for rapid and high-resolution pavement monitoring. Recent studies have shown that UAV imagery, photogrammetry, and deep learning (DL) can support automated detection and assessment of road-surface defects. For example, Aburqaq et al. \cite{aburqaq2025} demonstrated the use of UAV-based photogrammetry for flexible pavement inspection, while Alzamzami et al. \cite{alzamzami2024} proposed PDS-UAV, a YOLOv8-based pothole detection system using UAV imagery. Large-scale datasets and benchmark studies, such as RDD2022 \cite{arya2024}, have further supported the development of DL models for road-damage detection across different countries, defect types, and environmental conditions. In parallel, lightweight models such as YOLOv8-PD \cite{zeng2024}, YOLO9tr \cite{youwai2024}, and MobiLiteNet \cite{hu2025} have demonstrated the feasibility of accurate and deployable road-distress monitoring. 

Simulation environments and digital twins (DTs) have been increasingly explored to reduce dependence on field data collection and enable controlled evaluation of road-inspection systems. Tsai et al. \cite{tsai2020} generated virtual pothole scenes under different surface conditions and combined synthetic with real data to enhance road-distress detection. Wang et al. \cite{wang2023} reconstructed UAV-derived pavement backgrounds and rendered multiple distress classes under challenging visual conditions, while Wang et al. \cite{wang2024} integrated textured pavement-background modelling with Unreal Engine-based virtual UAV inspection to improve detection performance through synthetic-data augmentation. Collectively, these studies highlight the value of simulation for road-distress data generation and detector training.

System-level UAV inspection and digital-twin frameworks have also been proposed for road monitoring, route planning, and maintenance support. Silva et al. \cite{silva2020} introduced a distributed multi-agent architecture for pothole recognition in UAV images, Zhao et al. \cite{zhao2023} combined UAV path planning with image splicing for pavement-damage detection, and Zhong et al. \cite{zhong2023} investigated UAV swarm route planning for road inspection. In infrastructure management, DTs have been used for pavement-health monitoring, predictive maintenance, and decision support \cite{sierra2022,talaghat2024,consilvio2023,topu2025}. However, existing system-level, route-planning, and digital-twin approaches remain largely focused on image acquisition, monitoring architectures, route optimisation, or asset-level maintenance. Fewer studies provide a unified traffic-aware simulation testbed that jointly evaluates UAV inspection missions, procedural road-defect generation, dynamic vehicle and pedestrian occlusions, in-flight detection, tracking, and recovery strategies before real-world deployment.

\section{Digital Twin Simulation Framework}
\label{sec:framework}

This work proposes a digital twin (DT) simulator designed to provide a controlled and reproducible environment for UAV-based pavement-inspection research. It represents a configurable urban road network and supports experimentation with computer vision and machine learning (ML) algorithms. The simulator enables data acquisition through two inspection platforms: a ground vehicle and a UAV, as shown in Figure \ref{fig:simulator-components}. The following sections describe its main components.

\begin{figure*}[htpb]
\centering

\begin{subfigure}[b]{0.31\textwidth}
\centering
\includegraphics[width=\textwidth]{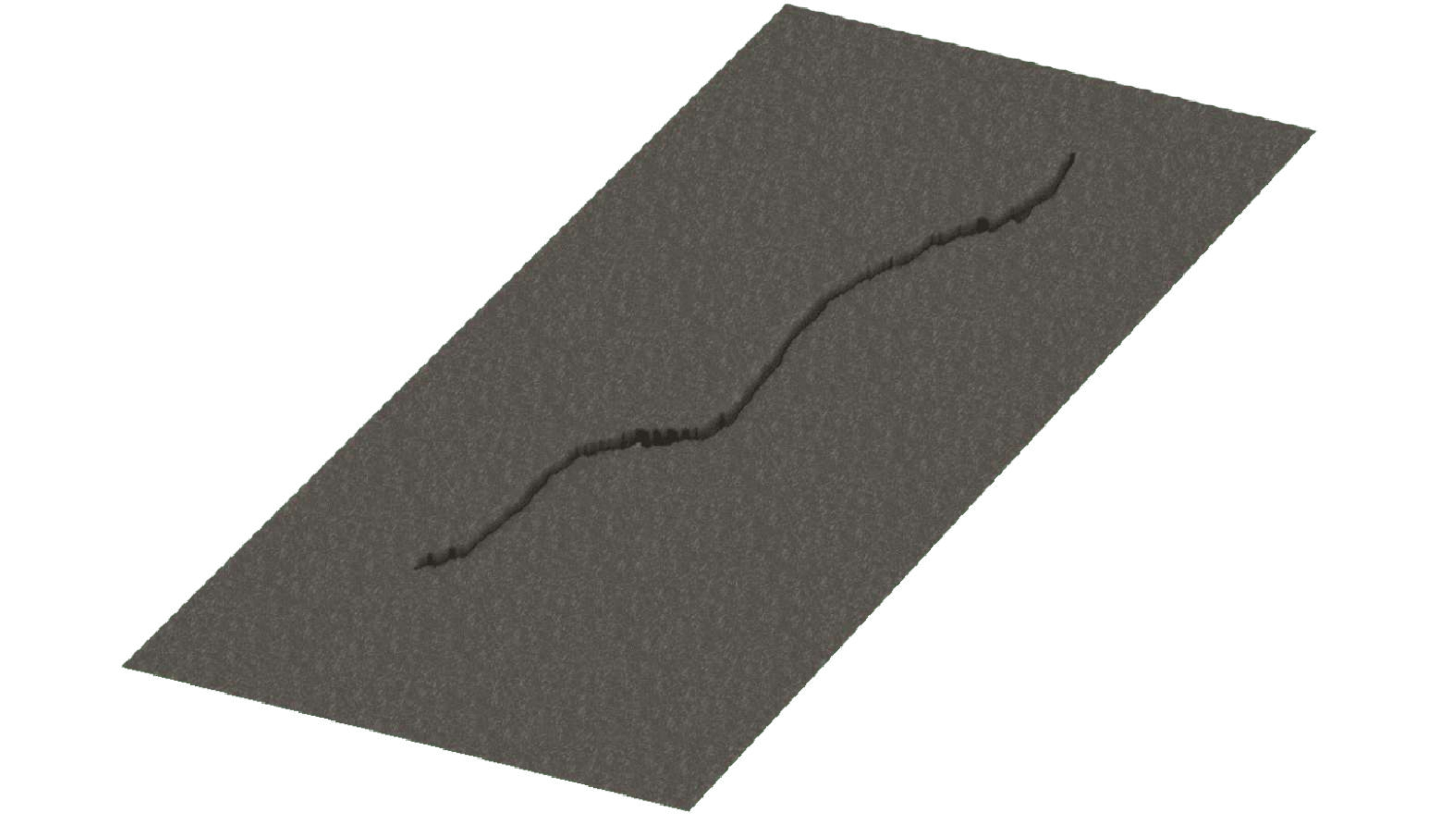}
\caption{}
\label{fig:crack-sample}
\end{subfigure}
\hfill
\begin{subfigure}[b]{0.31\textwidth}
\centering
\includegraphics[width=\textwidth]{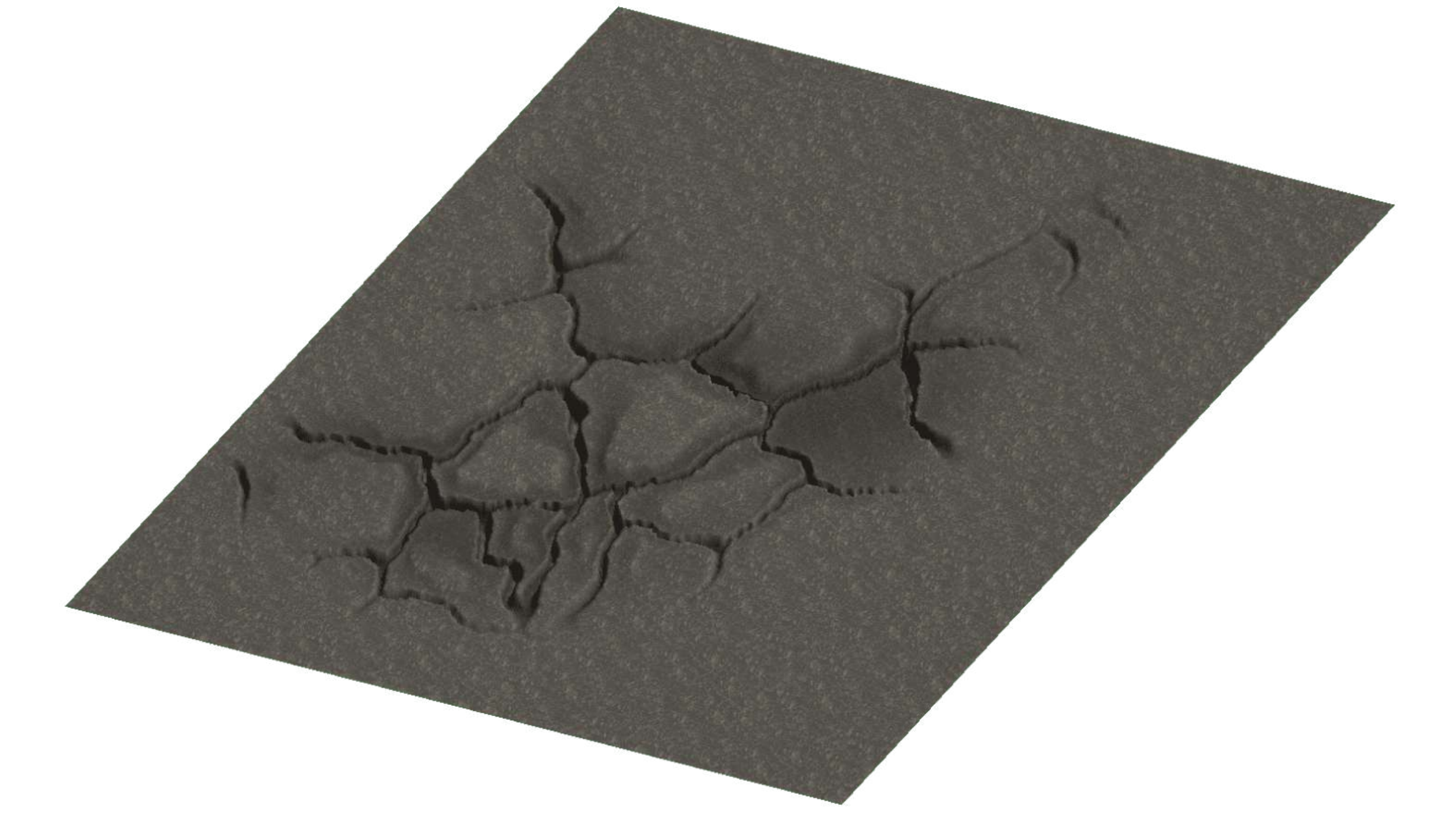}
\caption{}
\label{fig:crocodile-sample}
\end{subfigure}
\hfill
\begin{subfigure}[b]{0.31\textwidth}
\centering
\includegraphics[width=\textwidth]{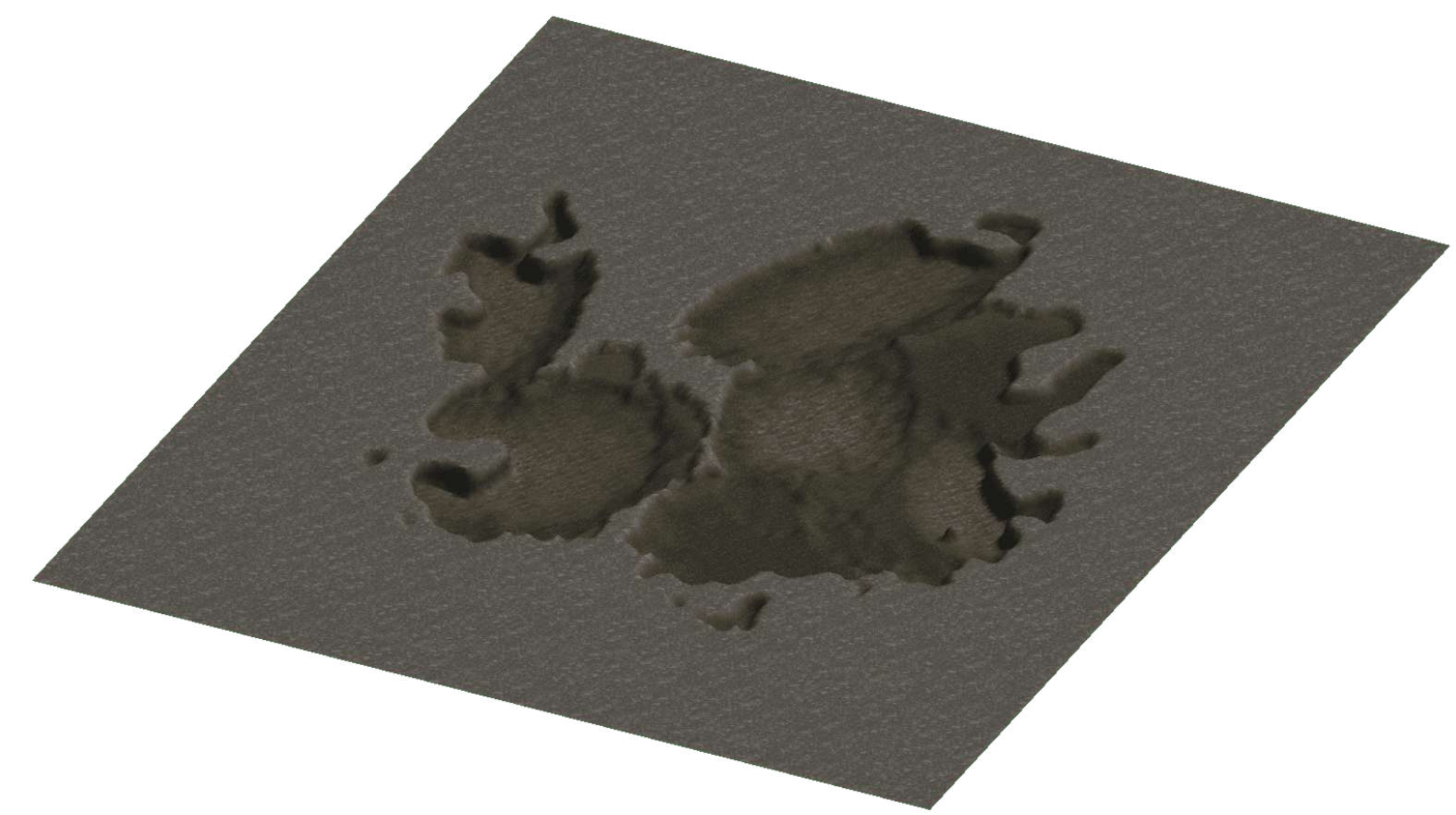}
\caption{}
\label{fig:pothole-sample}
\end{subfigure}

\caption{Isometric views of procedurally generated road-surface defects: (a) single crack, (b) crocodile crack, and (c) pothole.}
\label{fig:generated-defect-samples}
\end{figure*}

\subsection{Unity-Based Environment}

The virtual environment was developed in Unity, with selected assets modelled in Blender, to represent an urban road network composed of streets with varying widths, intersections, traffic lights, and road signs. Within this environment, the simulator includes a ground vehicle based on a 2018 Hyundai Grand i10 model and a UAV based on the DJI Phantom 4. The ground vehicle supports auxiliary data collection, although the experiments in this work focus on UAV-based inspection. To approximate realistic inspection conditions, vehicular traffic and pedestrian movement were implemented in Unity, allowing the UAV and ground vehicle to operate under dynamic urban conditions. A map of the environment with indicators for mobile elements is presented in Figure \ref{fig:simulator-map}.

Following the taxonomy used in \cite{salcedo2022}, road-surface deterioration can be described through three widely recognised defect categories. Cracks correspond to linear pavement fractures, while interconnected fracture patterns associated with advanced deterioration are represented as crocodile cracks. Potholes, in turn, represent localised bowl-shaped depressions caused by pavement material loss. Together, these categories represent increasing levels of road-surface deterioration. These three categories were procedurally generated and integrated into the virtual environment, as shown in Figure \ref{fig:generated-defect-samples}. The generated defects vary in shape, size, and edge irregularity to support the assessment of different road-damage conditions within the simulated inspection environment.

\subsection{NavMesh-Based UAV Route Planning and Inspection}

The UAV operates autonomously using Unity's NavMesh navigation system. In the simulator, an inspection target is represented as a set of road segments, where each segment $s$ is defined by start and end positions. Targets may be assigned interactively through the minimap interface or automatically by the experiment automator, which selects a randomised sequence of unvisited segments for repeatable batch evaluation. Once a target is assigned, the UAV computes a route from its position to the target segment using the baked NavMesh, which represents the traversable flight domain above the road network as a set of connected convex polygons. The navigation and inspection process consists of four stages: NavMesh generation through geometry baking, global pathfinding using Unity's internal navigation search, trajectory smoothing before execution by the \textit{NavMeshAgent}, and segment-level inspection using the selected navigation policy.

For reproducibility, the NavMesh configuration was defined using the UAV agent dimensions and motion constraints. The agent radius was set to 0.5 m, while the inspection altitude was configurable through the simulator interface and evaluated at three levels: 6 m, 10 m, and 15 m. The \textit{NavMeshAgent} was configured with a speed of 5 m/s, an angular speed of $45^\circ$/s, an acceleration of 8 m/s$^2$, and a stopping distance of 1 m. These parameters influence route feasibility, turning behaviour, mission time, and safe clearance during inspection. Dynamic avoidance was handled through the \textit{NavMeshAgent} local-avoidance settings, allowing the UAV to react to nearby agents or obstacles while following the global path computed by Unity's navigation system.

After reaching the assigned road segment, the UAV follows the configured inspection pattern while the DT monitors visibility, coverage, and recovery events. When an occlusion occurs, the UAV applies the recovery strategy defined in Section \ref{sec:digital-twin}. Road-defect outputs from the perception module described in Section \ref{sec:road-damage-detection} trigger automatic image capture and georeferencing, with samples saved only when defects are sufficiently centred in the UAV camera view. The simulator also records coverage, recovery ratio, mission time, and energy consumption for each segment, enabling the recovery strategies to be compared across traffic densities and flight altitudes. Once a segment has been processed, the UAV either proceeds to the next assigned segment or returns to its take-off position at the end of the mission.

\subsection{Digital Twin Definition}
\label{sec:digital-twin}

In this work, the term digital twin refers to a simulation-based operational replica of the UAV pavement-inspection process, designed to support controlled evaluation of inspection policies before field deployment. The DT models the states of the road network, the UAV, and dynamic traffic agents, which are continuously updated within the simulated environment to support online decision-making under occlusion and safety constraints. Unlike a static simulator, the proposed DT enables the evaluation and selection of recovery actions during mission execution. Formally, at discrete time $t$, the DT is represented as:

\begin{equation}
\mathcal{D}_t = \big( \mathcal{S}, \mathcal{A}_t, x_t, \mathcal{M}_t, \Pi \big),
\end{equation}
where $\mathcal{S} = \{s_1, s_2, \dots, s_N\}$ denotes the set of road segments in the road network, $\mathcal{A}_t$ the set of dynamic agents, $x_t$ the UAV state, $\mathcal{M}_t$ the inspection memory, and $\Pi$ the set of candidate recovery policies. An example road segment $s$ is shown in green in Figure \ref{fig:simulator-map}.

The UAV state at time $t$ is defined as:
\begin{equation}
x_t = (p_t, h_t, v_t, e_t),
\label{eq:uav-state}
\end{equation}
where $p_t \in \mathbb{R}^2$ is the planar position, $h_t \in \mathbb{R}_{>0}$ is the altitude, $v_t \in \mathbb{R}_{\ge 0}$ is the speed, and $e_t \in [0,1]$ is the battery level. Each dynamic agent $a_i \in \mathcal{A}_t$ is represented as:
\begin{equation}
a_i^t = (y_i^t, \nu_i^t, \kappa_i),
\label{eq:agent}
\end{equation}
where $y_i^t$ is the position, $\nu_i^t$ is the velocity, and $\kappa_i \in \{\text{vehicle}, \text{pedestrian}\}$ is the class label.

For each road segment $s$, we define a visibility score $o_{s,t} \in [0,1]$, where $o_{s,t}=1$ indicates that the segment is fully observable by the UAV at time $t$, and $o_{s,t}=0$ indicates that it is completely occluded or outside the field of view. Intermediate values represent partial visibility due to dynamic occlusions. The inspection memory is then defined as:
\begin{equation}
\mathcal{M}_t = \{m_s^t \mid s \in \mathcal{S}\},
\end{equation}
where each segment state is given by:
\begin{equation}
m_s^t =
\begin{cases}
\texttt{inspected}, & \text{if } o_{s,t} \ge \tau_o,\\
\texttt{pending}, & \text{otherwise.}
\end{cases}
\end{equation}

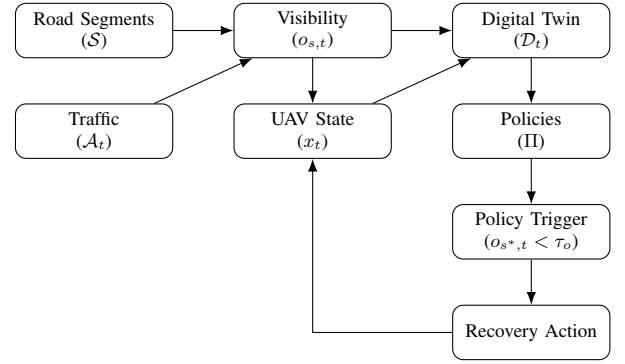
\begin{figure}[t]
\centering
\begin{tikzpicture}[
    scale=0.80,
    transform shape,
    node distance=0.75cm and 1cm,
    box/.style={
        draw,
        rectangle,
        rounded corners,
        align=center,
        minimum width=2.6cm,
        minimum height=0.9cm,
        font=\small
    },
    >=Latex
]

\node[box] (env) {Road Segments\\($\mathcal{S}$)};
\node[box, below=of env] (traffic) {Traffic\\($\mathcal{A}_t$)};

\node[box, right=of env] (vis) {Visibility\\($o_{s,t}$)};
\node[box, below=of vis] (uav) {UAV State\\($x_t$)};

\node[box, right=of vis] (dt) {Digital Twin\\($\mathcal{D}_t$)};
\node[box, below=of dt] (policy) {Policies\\($\Pi$)};
\node[box, below=of policy] (trigger) {Policy Trigger\\($o_{s^\ast,t}<\tau_o$)};
\node[box, below=of trigger] (action) {Recovery Action};

\draw[->] (env) -- (vis);
\draw[->] (traffic) -- (vis);

\draw[->] (vis) -- (uav);

\draw[->] (vis) -- (dt);
\draw[->] (uav) -- (dt);

\draw[->] (dt) -- (policy);
\draw[->] (policy) -- (trigger);
\draw[->] (trigger) -- (action);

\draw[->] (action.west) -| (uav.south);

\end{tikzpicture}
\caption{Digital twin architecture for traffic-aware UAV pavement monitoring. The digital twin integrates road segments, traffic agents, UAV state, and visibility estimates to trigger recovery actions under dynamic occlusions.}
\label{fig:digital_twin_architecture}
\end{figure}

When the visibility of the currently targeted segment $s^\ast$ falls below a predefined threshold, i.e., $o_{s^\ast,t} < \tau_o$, the DT activates a recovery policy selected from $\Pi$. The Baseline strategy follows the planned inspection route without applying any recovery action when occlusions occur. The proposed framework considers three policies to handle occlusions: hover-and-recheck, micro-repositioning, and skip-and-revisit, as summarised in Table \ref{tab:policies}. The waiting time $\Delta t_{\text{wait}}$ represents a short duration during which the UAV remains stationary, while the displacement vector $\delta_t \in \mathbb{R}^2$ defines a local movement applied during the micro-repositioning strategy, constrained by $\|\delta_t\| \le d_{\max}$, where $d_{\max}$ is the maximum allowed displacement. Figure \ref{fig:digital_twin_architecture} illustrates the overall DT architecture.

\begin{table}[t]
\centering
\caption{Adaptive recovery policies.}
\label{tab:policies}
\begin{tabular}{m{1cm}|m{1.8cm}|m{3.6cm}}
\hline
\multicolumn{1}{c|}{\textbf{Policy}} &
\multicolumn{1}{c|}{\textbf{Action}} &
\multicolumn{1}{c}{\textbf{Description}} \\
\hline

Hover & $p_{t+1} = p_t$ &
Remains stationary for a short duration $\Delta t_{\text{wait}}$ to allow temporary occlusions to clear. \\

Micro & $p_{t+1} = p_t + \delta_t$ &
Performs a small displacement ($\|\delta_t\| \le d_{\max}$) to improve visibility while remaining close to the target segment. \\

Skip & \makecell{Continue to\\next segment} &
Skips the occluded segment and revisits it later following the predefined inspection path. \\

\hline
\end{tabular}
\end{table}

\begin{table*}[htpb]
\centering
\caption{Summary of the datasets used to construct the final five-class dataset.}
\label{tab:dataset-summary}
\resizebox{\textwidth}{!}{%
\begin{tabular}{lrrrrrrrrr}
\toprule
 & \multicolumn{6}{c}{\textbf{Instances per class}} & \multicolumn{3}{c}{\textbf{Images}} \\
\cmidrule(lr){2-7} \cmidrule(lr){8-10}
\textbf{Source Dataset} &
\textbf{Single} &
\textbf{Crocodile} &
\textbf{Pothole} &
\textbf{Person} &
\textbf{Car} &
\textbf{Total} &
\textbf{Annotated} &
\textbf{Background} &
\textbf{Total} \\
\midrule
HighRPD \cite{highrpd}                  
& 11{,}409 & 6{,}900 & 0 & 0 & 0 & 18{,}309 & 9{,}974 & 310 & 10{,}284 \\

Pothole-Recog. \cite{silva2020}         
& 108 & 0 & 453 & 0 & 0 & 561 & 111 & 11 & 122 \\

PothRGBD \cite{Yurdakul2025}            
& 0 & 0 & 972 & 0 & 0 & 972 & 871 & 0 & 871 \\

UAPD \cite{uapd,uapd2}                        
& 3{,}256 & 0 & 94 & 0 & 0 & 3{,}350 & 2{,}146 & 0 & 2{,}146 \\

UAV-PDD2023 \cite{uavpdd2023}           
& 10{,}074 & 603 & 195 & 0 & 0 & 10{,}872 & 2{,}403 & 0 & 2{,}403 \\

RDD2022\textsuperscript{*} \cite{arya2024}
& 2{,}689 & 293 & 86 & 0 & 0 & 3{,}068 & 1{,}919 & 482 & 2{,}401 \\

UAV car detection \cite{car_roboflow}   
& 0 & 0 & 0 & 0 & 16{,}868 & 16{,}868 & 299 & 0 & 299 \\

Pedestrian recognition \cite{pedestrian_roboflow}  
& 0 & 0 & 0 & 17{,}034 & 0 & 17{,}034 & 215 & 0 & 215 \\
\midrule
\textbf{Merged Dataset} 
& 27{,}536 & 7{,}796 & 1{,}800 & 17{,}034 & 16{,}868 & 71{,}034 & 17{,}938 & 803 & 18{,}741 \\

\textbf{Balanced Dataset} 
& 24{,}295 & 24{,}295 & 24{,}295 & 24{,}000 & 23{,}884 & 120{,}769 & 42{,}755 & 3{,}420 & 46{,}175 \\

\textbf{Synthetic Dataset} 
& 4{,}435 & 3{,}134 & 4{,}665 & 8{,}376 & 5{,}333 & 25{,}943 & 2{,}235 & 0 & 2{,}235 \\
\bottomrule
\end{tabular}%
}
\vspace{0.3em}
\begin{minipage}{\textwidth}
\footnotesize
\textsuperscript{*}Only the China UAV subset of RDD2022 was considered.
\end{minipage}
\end{table*}

\section{Multitask UAV Perception Model}
\label{sec:road-damage-detection}

The proposed framework embeds a perception layer within the UAV platform, which observes the environment from a top-down perspective. During UAV inspection, the detector processes RGB frames from the simulated camera to identify road-surface defects, pedestrians, and vehicles. Road-defect detections support automatic image capture and dataset generation, while pedestrian and vehicle detections support traffic-aware inspection by identifying objects that may occlude the road surface. In this way, the detection module links the visual simulation environment with the DT decision layer described in Section \ref{sec:digital-twin}. The YOLO family of object-detection models was selected because of its balance between inference speed and detection accuracy. 

\subsection{Dataset Description and Preprocessing}
\label{subsec:dataset-prep}

A multi-source dataset was constructed from six road-damage datasets and two additional UAV datasets for traffic-agent classes. Since these sources differed in annotation format, image resolution, and original class taxonomy, they were normalised into a common three-class road-defect scheme consisting of \textit{Single Crack}, \textit{Crocodile Crack}, and \textit{Pothole}. HighRPD \cite{highrpd} originally provided three classes: \textit{line}, \textit{block}, and \textit{pit}. The \textit{line} and \textit{block} classes were remapped to \textit{Single Crack} and \textit{Crocodile Crack}, respectively, while images containing at least one \textit{pit} annotation were discarded. The pothole-recognition dataset of Silva et al. \cite{silva2020} distinguished between \textit{crack} and \textit{pothole}, which were mapped directly to \textit{Single Crack} and \textit{Pothole}. PothRGBD \cite{Yurdakul2025} contained only pothole instances annotated as polygons; these annotations were converted to enclosing bounding boxes and retained under the \textit{Pothole} class.

UAPD \cite{uapd}, \cite{uapd2} and UAV-PDD2023 \cite{uavpdd2023} followed a common distress taxonomy including longitudinal, transverse, oblique, and alligator cracks, as well as \textit{pothole} and the non-target \textit{repair} class. For both, alligator-pattern cracks were mapped to \textit{Crocodile Crack}, longitudinal, transverse, and oblique cracks were merged into \textit{Single Crack}, and \textit{repair} instances were discarded. Finally, the China UAV subset of RDD2022 \cite{arya2024} originally used D00, D10, D20, and D40 labels for longitudinal cracks, transverse cracks, alligator cracks, and potholes, respectively, together with \textit{repair} and \textit{block} labels. D00 and D10 were merged into \textit{Single Crack}, D20 was mapped to \textit{Crocodile Crack}, D40 was mapped to \textit{Pothole}, and the \textit{repair} and \textit{block} labels were discarded.

After class-level normalisation, all annotations were converted into a common Pascal VOC/XML representation to enable consistent merging across sources. The unified dataset was then exported into the YOLO object-detection format for model training. Images were downscaled to a maximum side length of 640 px, with bounding-box coordinates updated accordingly. Images without valid annotations were discarded unless they were later used as verified background samples. Table \ref{tab:dataset-summary} reports the resulting object-instance distribution. Because the merged dataset was strongly imbalanced across defect classes, an augmentation pipeline was applied using the Albumentations library \cite{buslaev2020}. The transformations included horizontal flipping, brightness and contrast adjustment, Gaussian noise, and rotations of up to $\pm15^{\circ}$. Bounding boxes were updated during augmentation, and boxes with insufficient visible area after transformation were removed.

To make the detector suitable for traffic-aware inspection, two additional classes were incorporated: \textit{Person} and \textit{Car}. Object instances for both categories were extracted from public top-view UAV datasets \cite{pedestrian_roboflow,car_roboflow} and composited onto the merged road-damage images under controlled placement rules. A manually reviewed background set was also included to reduce false positives on undamaged road surfaces. The background images were verified to contain no visible road damage, people, or vehicles before being resized and augmented using a lighter transformation pipeline. As shown in Table \ref{tab:dataset-summary}, the final \textit{Balanced Dataset} contains 46,175 images and 120,769 annotated bounding boxes across the five classes. The \textit{Balanced Dataset} was split into training, validation, and test subsets using a class-stratified 70\%/20\%/10\% ratio, ensuring that the relative proportion of each class was preserved across all three subsets.

Finally, since the target domain is a Unity-based environment with synthetically generated road damage, vehicles, and pedestrians, we collected an additional \textit{Synthetic Dataset}. This dataset consists of top-view UAV captures collected within the DT simulation framework, showing road scenes with instances from the five classes.

\subsection{Model Development and Training}
\label{sec:training}

To select the object detector with the best trade-off between accuracy, speed, and deployment cost for simulator integration, a preliminary first-stage detection experiment was conducted. YOLOv8 \cite{Ultralytics2023}, YOLO11 \cite{ultralytics2024yolov11}, and YOLO12 \cite{tian2025} were compared across their nano, small, medium, and large variants using three coarse detection classes: \textit{Road Defect}, \textit{Person}, and \textit{Car}. For this model-selection experiment, the road-defect subtypes \textit{Single Crack}, \textit{Crocodile Crack}, and \textit{Pothole} were merged into a single \textit{Road Defect} class, while \textit{Person} and \textit{Car} were retained as separate classes. All models were initialised from COCO pre-trained weights and evaluated on the same dataset split. The evaluation considered detection-accuracy and computational-efficiency metrics, including mean average precision (mAP) at an intersection-over-union (IoU) threshold of 0.5, mAP averaged over IoU thresholds from 0.5 to 0.95, precision, recall, model size, frames per second (FPS), and giga floating-point operations (GFLOPs), following the definitions in \cite{salcedo2024}. Based on the results reported in Table \ref{tab:yolo-models-comparison}, YOLOv8n was selected as the most suitable trade-off between detection accuracy, model size, computational cost, and inference speed.

After selecting YOLOv8n as the base detector under the three-class configuration used during model selection, we implemented a modified YOLOv8n architecture capable of producing the final five-class output required by the simulator. In the first-stage detector, the road-defect categories were treated as a single \textit{Road Defect} class, alongside \textit{Person} and \textit{Car}, preserving the same detection setup used for comparing candidate YOLO models. To recover the fine-grained road-defect labels, a lightweight classification head was then integrated using road-defect feature-pyramid embeddings to distinguish among \textit{Single Crack}, \textit{Crocodile Crack}, and \textit{Pothole}. In this way, the final system retained the efficient detection capability of YOLOv8n while extending it to provide the five semantic classes required for simulator-based evaluation.

\begin{figure*}[ht]
    \centering
    \includegraphics[width=\linewidth]{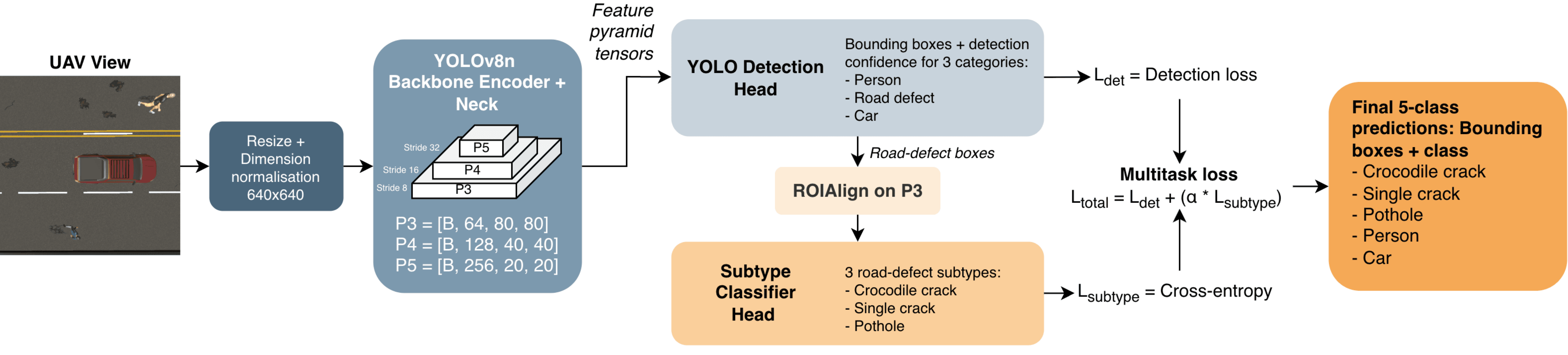}
    \caption{Proposed multitask model based on YOLOv8n. B and P denote batch size and pyramid level, respectively.}
    \label{fig:model-architecture}
\end{figure*}

The YOLOv8n encoder and both heads were first trained using the \textit{Balanced Dataset}, detailed in Table \ref{tab:dataset-summary}, which contains real-world image samples. To better adapt the model to the target domain, the trained shared-backbone YOLOv8n multitask model was used as the initialisation checkpoint and further optimised on the \textit{Synthetic Dataset}. As illustrated in Figure \ref{fig:model-architecture}, the multitask model first detects road defects and dynamic occluding agents, while the second head performs fine-grained classification only on regions identified as road damage. Thus, the YOLO backbone is used once to encode each image, and only embeddings corresponding to detected road defects are passed to the second head, enabling efficient road-distress classification.

In the implemented architecture, the YOLOv8n backbone and neck produce a feature pyramid composed of the P3, P4, and P5 tensors. These tensors are used by the YOLO detection head to predict coarse object categories: \textit{Road Defect}, \textit{Person}, and \textit{Car}. For the subtype branch, only the P3 feature map is used, since it has the highest spatial resolution in the feature pyramid and is therefore better suited for representing small or thin road-surface defects. Road-defect bounding boxes are projected onto this P3 feature map and processed using ROIAlign, which extracts fixed-size feature representations for each road-defect region without cropping the original RGB image or executing a second backbone. These ROI-aligned features are then passed to the subtype classifier to distinguish \textit{Crocodile Crack}, \textit{Single Crack}, and \textit{Pothole}. The multitask objective combines the YOLO detection loss with the subtype classification loss as
\begin{equation}
L_{\mathrm{total}} = L_{\mathrm{det}} + \alpha L_{\mathrm{subtype}},
\end{equation}
where \(L_{\mathrm{det}}\) is the standard YOLO detection loss, \(L_{\mathrm{subtype}}\) is a cross-entropy loss over the three road-defect subtypes, and \(\alpha\) controls the contribution of the subtype branch. In this work, \(\alpha\) was set to 1.0, giving equal weight to the detection and subtype objectives.

\section{Experimental Results}
\label{sec:results}

This section reports the results of the DT recovery strategies and the UAV perception pipeline. The DT simulations were tested on a local desktop computer equipped with an AMD Ryzen 5 5600G processor at 3.90 GHz, an NVIDIA GeForce GTX 1050 Ti GPU with 4 GB of VRAM, 16 GB of RAM at 2666 MT/s, and a 64-bit x64 system architecture. The YOLO variants were trained under identical conditions, and FPS was measured on a single NVIDIA T4 GPU for consistency.

\subsection{Digital Twin Recovery Strategies} 

Table \ref{tab:main_results} reports the inspection coverage achieved by the four navigation strategies under different traffic densities and flight altitudes. Values are presented as the mean $\pm$ standard deviation over 20 independent experimental repetitions. Additional operational metrics, including mission time, energy consumption, and recovery ratio, were also logged and are discussed where they affect the interpretation of each strategy. Under low-traffic conditions, Baseline achieved the highest coverage at medium and high altitudes, reaching $84.01 \pm 5.72$\% and $82.82 \pm 6.79$\%, respectively. At low altitude, however, Micro produced the highest coverage ($60.30 \pm 11.81$\%). Although Skip enabled revisits to previously occluded regions, these additional manoeuvres did not translate into higher coverage in this configuration and instead increased mission duration and energy consumption.

\begin{table}[htpb]
\centering
\caption{Inspection coverage of the baseline and recovery strategies under different traffic densities and flight altitudes (mean $\pm$ standard deviation, \%). Low, medium, and high altitudes correspond to 6 m, 10 m, and 15 m, respectively.}
\label{tab:main_results}
\resizebox{\columnwidth}{!}{
\begin{tabular}{llcccc}
\toprule
\textbf{Traffic} & \textbf{Altitude} & \textbf{Baseline} & \textbf{Hover} & \textbf{Micro} & \textbf{Skip} \\
\midrule
Low    & Low    & $56.00 \pm 11.48$ & $52.35 \pm 11.89$ & $60.30 \pm 11.81$ & $47.45 \pm 11.78$ \\
Low    & Medium & $84.01 \pm 5.72$  & $83.59 \pm 7.74$  & $82.70 \pm 10.47$ & $79.63 \pm 10.64$ \\
Low    & High   & $82.82 \pm 6.79$  & $82.45 \pm 10.29$ & $76.41 \pm 8.83$  & $77.75 \pm 8.12$ \\
Medium & Low    & $52.37 \pm 10.18$ & $53.90 \pm 11.55$ & $53.58 \pm 9.26$  & $51.71 \pm 11.81$ \\
Medium & Medium & $84.53 \pm 8.15$  & $80.43 \pm 11.01$ & $81.82 \pm 7.35$  & $81.03 \pm 9.10$ \\
Medium & High   & $75.19 \pm 11.42$ & $81.98 \pm 9.99$  & $74.60 \pm 10.70$ & $75.03 \pm 11.52$ \\
High   & Low    & $50.78 \pm 10.74$ & $54.79 \pm 14.08$ & $51.80 \pm 11.40$ & $49.45 \pm 11.22$ \\
High   & Medium & $81.87 \pm 11.16$ & $81.51 \pm 7.98$  & $79.13 \pm 9.61$  & $74.91 \pm 12.30$ \\
High   & High   & $78.28 \pm 10.47$ & $75.53 \pm 9.85$  & $73.01 \pm 15.55$ & $76.84 \pm 11.03$ \\
\bottomrule
\end{tabular}
}
\end{table}

\begin{table*}[ht]
\centering
\caption{Comparison of YOLOv8, YOLO11, and YOLO12 variants for first-stage detection.}
\label{tab:yolo-models-comparison}
\resizebox{\textwidth}{!}{
\begin{tabular}{lcccccccccccc}
\toprule
\multirow{2}{*}{\textbf{Metrics}} &
\multicolumn{4}{c}{\textbf{YOLOv8}} &
\multicolumn{4}{c}{\textbf{YOLO11}} &
\multicolumn{4}{c}{\textbf{YOLO12}} \\
\cmidrule(lr){2-5} \cmidrule(lr){6-9} \cmidrule(lr){10-13}
& n & s & m & l & n & s & m & l & n & s & m & l \\
\midrule
mAP@0.5 & 0.866 & 0.875 & 0.878 & 0.876 & 0.869 & 0.876 & 0.874 & 0.873 & 0.874 & 0.875 & 0.872 & 0.869 \\
mAP@0.50:0.95 & 0.775 & 0.785 & 0.791 & 0.789 & 0.778 & 0.785 & 0.785 & 0.783 & 0.784 & 0.786 & 0.783 & 0.780 \\
Precision & 0.888 & 0.887 & 0.905 & 0.898 & 0.890 & 0.895 & 0.893 & 0.889 & 0.901 & 0.890 & 0.896 & 0.885 \\
Recall & 0.840 & 0.856 & 0.849 & 0.855 & 0.847 & 0.853 & 0.851 & 0.854 & 0.844 & 0.854 & 0.847 & 0.847 \\
Size (MB) & 6.0 & 21.5 & 49.6 & 83.6 & 5.2 & 18.3 & 38.6 & 48.8 & 5.3 & 18.1 & 38.9 & 51.0 \\
FPS & 273.47 & 97.73 & 39.19 & 22.68 & 263.11 & 92.38 & 35.98 & 32.36 & 166.43 & 63.46 & 26.03 & 18.18 \\
GFLOPs & 8.1 & 28.4 & 78.7 & 164.8 & 6.3 & 21.3 & 67.7 & 86.6 & 6.3 & 21.2 & 67.1 & 88.5 \\
\bottomrule
\end{tabular}
}
\end{table*}

Under medium-traffic conditions, Hover achieved the highest coverage at low and high altitudes ($53.90 \pm 11.55$\% and $81.98 \pm 9.99$\%), whereas Baseline performed best at medium altitude ($84.53 \pm 8.15$\%). The differences between Hover and Baseline were relatively small, especially at medium and high altitudes, indicating that the benefit of recovery depends on the local traffic and flight configuration. Similarly, Skip continued to trigger revisit actions, but these actions mainly increased mission time and energy use without producing a substantial coverage gain. Micro remained broadly stable across the evaluated settings, although its coverage was generally lower than that of Baseline or Hover.

Under high-traffic conditions, low-altitude coverage decreased for all strategies because persistent vehicle occlusions reduced the visible road surface. Hover achieved the highest mean coverage at low altitude ($54.79 \pm 14.08$\%), but the large standard deviation indicates considerable run-to-run variability. In contrast, Baseline achieved the highest coverage at medium altitude ($81.87 \pm 11.16$\%) and high altitude ($78.28 \pm 10.47$\%). The clearest operational cost was observed for Hover under high traffic and low altitude, where the average mission time increased to $91.40 \pm 118.27$~s and energy consumption reached $5.08 \pm 6.57$\%. This occurred because Hover waits for the occluding vehicle to leave the inspection area before continuing; during congested runs, vehicles sometimes remained stationary for long periods, forcing the UAV to wait over the same location. Overall, the recovery strategies introduced trade-offs between coverage, recovery capability, mission duration, and energy consumption, but they did not consistently outperform Baseline in terms of coverage under the evaluated inspection configuration.

\subsection{Multitask UAV Perception Model}

Table \ref{tab:yolo-models-comparison} presents the preliminary comparison of YOLOv8, YOLO11, and YOLO12 across their nano, small, medium, and large variants for first-stage detection, considering three coarse classes: \textit{Road Defect}, \textit{Person}, and \textit{Car}. The results show that detection accuracy does not improve substantially when moving from nano to larger variants. For example, mAP@0.5 increases only from 0.866 for YOLOv8n to 0.878 for YOLOv8m, while inference speed drops from 273.47 FPS to 39.19 FPS and model size increases from 6.0 MB to 49.6 MB. A similar pattern is observed for YOLO11 and YOLO12. Therefore, YOLOv8n was selected due to its favourable balance between accuracy, training efficiency, model size, and real-time inference speed. These characteristics make it suitable for deployment inside the Unity simulator, where inference must run in real time.

\begin{figure}[h] 
\centering 
\includegraphics[width=0.48\textwidth]{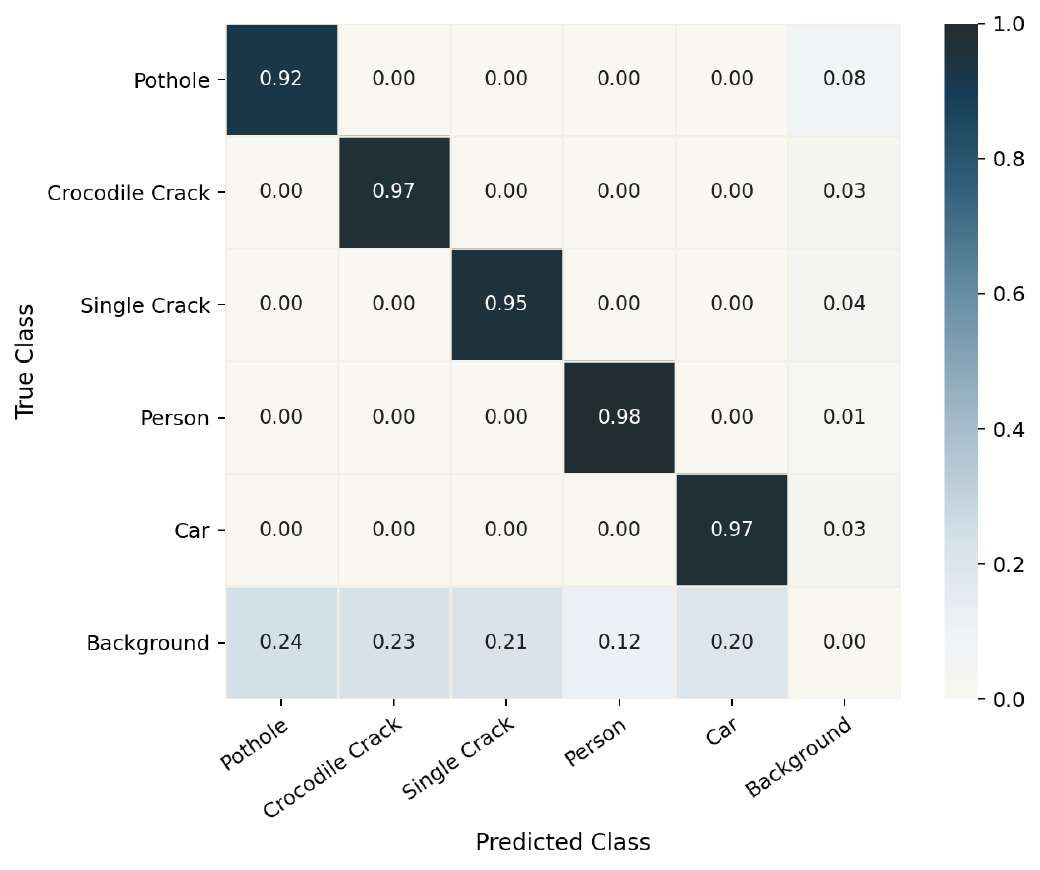} \caption{Confusion matrix of the multitask detection model, fine-tuned and evaluated on the \textit{Synthetic Dataset}.} 
\label{fig:cm} 
\end{figure}

The YOLOv8n-based multitask model described in Section \ref{sec:training} was initially trained using the \textit{Balanced Dataset} and subsequently fine-tuned with the \textit{Synthetic Dataset}, as it represents the target domain. The first-stage detector was trained for 40 epochs, while the second-stage classifier was trained for 10 epochs, both using early stopping and learning-rate scheduling. Evaluation was performed on the simulator test split and covered all five classes required by the DT. Before synthetic fine-tuning, the multitask YOLOv8n model achieved a five-class mAP@0.5 of 0.7442, mAP@0.50:0.95 of 0.6124, and macro F1-score of 0.7666. After fine-tuning, performance improved substantially, reaching 0.9591 mAP@0.5, 0.6450 mAP@0.50:0.95, and 0.9400 macro F1-score. The detection head also improved from 0.8525 to 0.9738 mAP@0.5, while the subtype branch remained highly reliable, reaching a macro F1-score of 0.9973 on matched predicted road-defect ROIs. The confusion matrix shown in Figure \ref{fig:cm} further indicates that synthetic-domain fine-tuning improved localisation and detection coverage.

\section{Conclusions}
\label{sec:conclusions}

This paper presented a Unity-based digital twin framework for traffic-aware UAV pavement monitoring in open-traffic conditions. The framework integrates procedurally generated road defects, vehicles and pedestrians, UAV navigation, road-damage detection, and adaptive recovery strategies for occluded road regions. A multitask YOLOv8n perception model was integrated into the simulator to detect road defects and traffic agents while classifying road-defect subtypes using ROI-aligned features from the same image encoding. After fine-tuning on the synthetic simulator dataset, the model achieved 0.9591 mAP@0.5, 0.6450 mAP@0.50:0.95, and 0.9400 macro F1-score for the final five-class output, indicating that synthetic-domain adaptation substantially improved the perception module used by the digital twin.

The simulation experiments showed that flight altitude had a strong effect on inspection coverage and that no single recovery strategy dominated across all traffic conditions. The evaluated policies produced different trade-offs: hover-and-recheck improved coverage in some medium- and high-traffic settings but could increase mission time under congestion, while skip-and-revisit produced non-zero recovery ratios at the cost of longer missions and higher energy consumption. Future work will focus on validating the framework with real UAV imagery, extending the traffic conditions considered in the simulator, incorporating weather effects, and developing adaptive policy-selection methods that choose recovery actions according to the current occlusion, altitude, and traffic state.

\section*{Author Contributions}
Conceptualization, Y.U. and E.S.; methodology, Y.U., G.L., E.S., and M.F.; software, Y.U., G.L., E.S., and M.F.; validation, Y.U. and E.S.; formal analysis, Y.U., E.S., and G.L.; investigation, Y.U., G.L., and E.S.; data curation, G.L.; writing---original draft preparation, Y.U., G.L., and E.S.; writing---review and editing, E.S.; visualization, Y.U., G.L., and E.S.; supervision, E.S. All authors have read and agreed to the published version of the manuscript.

\bibliographystyle{IEEEtran}
\bibliography{bibliography}

\end{document}